\pdfoutput=1

\documentclass[11pt]{article}

\usepackage{latex/acl}

\usepackage{times}
\usepackage{latexsym}
\usepackage{hyperref}
\usepackage[normalem]{ulem}
\usepackage[T1]{fontenc}

\usepackage[utf8]{inputenc}

\usepackage{microtype}

\usepackage{inconsolata}

\usepackage{times}  
\usepackage{algorithm}
\usepackage{algorithmic}
\usepackage{ulem}

\newcommand{\mname}{CaQR}

\newcommand{\strc}{CaQR(S)}
\newcommand{\local}{CaQR(R)}
\newcommand{\both}{CaQR}

\usepackage{xcolor}

\usepackage{subcaption}
\usepackage{amsmath,amssymb,amsfonts, amsbsy}

\usepackage[export]{adjustbox}
\usepackage{multirow}
\usepackage{bm}
\usepackage{authblk}
\usepackage[utf8]{inputenc}

\usepackage{newfloat}
\usepackage{listings}
\DeclareCaptionStyle{ruled}{labelfont=normalfont,labelsep=colon,strut=off} 
\floatstyle{ruled}
\newfloat{listing}{tb}{lst}{}
\floatname{listing}{Listing}
\title{Improving Multi-hop Logical Reasoning in Knowledge Graphs with Context-Aware Query Representation Learning}

\author[1]{Jeonghoon Kim}
\author[1]{Heesoo Jung}
\author[2]{Hyeju Jang}
\author[1]{Hogun Park \thanks{Corresponding Author}}

\affil[1]{Sungkyunkwan University, Republic of Korea}
\affil[2]{Indiana University Indianapolis, USA}

\affil[ ]{\texttt{\{kjh9503, steve305\}@skku.edu}, \texttt{hyejuj@iu.edu}, \texttt{hogunpark@skku.edu}}

\begin{document}

\maketitle

\begin{abstract}\label{abstract}
Multi-hop logical reasoning on knowledge graphs is a pivotal task in natural language processing, with numerous approaches aiming to answer First-Order Logic (FOL) queries. Recent geometry (e.g., box, cone) and probability (e.g., beta distribution)-based methodologies have effectively addressed complex FOL queries. However, a common challenge across these methods lies in determining accurate geometric bounds or probability parameters for these queries. The challenge arises because existing methods rely on linear sequential operations within their computation graphs, overlooking the logical structure of the query and the relation-induced information that can be gleaned from the relations of the query, which we call the \textit{context of the query}. To address the problem, we propose a model-agnostic methodology that enhances the effectiveness of existing multi-hop logical reasoning approaches by fully integrating the context of the FOL query graph. Our approach distinctively discerns (1) the structural context inherent to the query structure and (2) the relation-induced context unique to each node in the query graph as delineated in the corresponding knowledge graph. This dual-context paradigm helps nodes within a query graph attain refined internal representations throughout the multi-hop reasoning steps. Through experiments on two datasets, our method consistently enhances the three multi-hop reasoning foundation models, achieving performance improvements of up to 19.5\%. Our code is available at \url{https://github.com/kjh9503/caqr}.
\end{abstract}

\section{Introduction}\label{introduction}

\begin{figure}[t]
    \centering
    \includegraphics[width=\columnwidth]{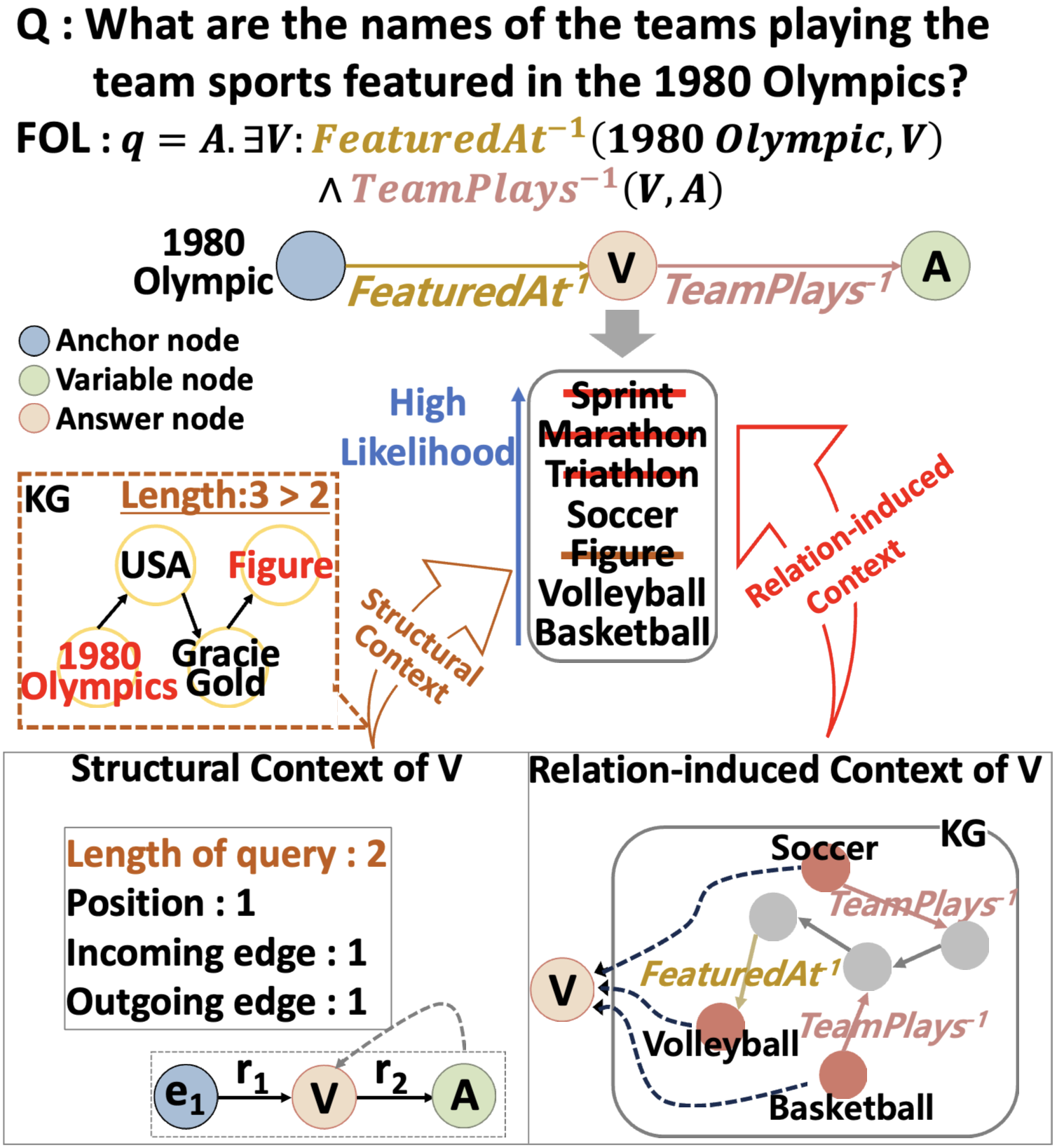}
    \caption{
    The existing methods may include wrong answers such as, \textit{Sprint}, \textit{Marathon}, \textit{Triathlon}, and \textit{Figure} because the candidates held by the variable node (V) in the inference process are only influenced by the \textit{1980 Olympic} and $FeaturedAt^{-1}$. However, our approach uses structural and relation-induced contexts to find a more accurate embedding of V, which helps us to predict answers that are close to the ground truth.
    }
    \vspace{-4mm}
    \label{fig:figure1}
\end{figure}


Multi-hop logical reasoning on Knowledge Graphs (KGs) is a crucial task in natural language processing. KGs, which map real-world knowledge as interconnected entities and relationships \cite{sinha2015overview, vrandevcic2014wikidata}, help answer complex questions expressed as First-Order Logic (FOL) queries. A FOL query can be converted into a computation graph, using variables, conjunctions, and existential quantification operators to represent a natural language question. For example, the computation graph of the query representing ``What are the names of the teams playing the team sports featured in the 1980 Olympics?'' can be constructed with two conjunction relations: 1) finding the sports featured in the 1980 Olympics, and 2) finding the sports teams that play that sports (see Figure 1 for illustration). 
Recent advancements in multi-hop reasoning leverage the power of embedding-based models such as Q2B \cite{q2b}, BetaE \cite{betae}, and ConE \cite{cone}. These models embed both a given query and entities (answer candidates) in a KG into a latent space. In this space, entities relevant to the query are positioned closer to the query's embedding. This allows the model to predict the answer to the query by identifying these nearby entities. 

Despite their success, these embedding-based models still face a major hurdle: learning more accurate geometric bounds or probability parameters for complex queries. Prior models handle all queries in the same way, building the embedding for each relationship (e.g., $FeaturedAt^{-1}$ or $TeamsPlay^{-1}$ in Figure \ref{fig:figure1}) one after another. This approach overlooks the structural context and relation-induced context within each question.
Therefore, entities like \textit{Sprint}, \textit{Marathon}, \textit{Triathlon}, and \textit{Figure} are treated as answers by the existing models~\cite{q2b, betae, cone} based on the given query in Figure~\ref{fig:figure1}.

To address this limitation, we propose a novel query embedding technique, named \textbf{CaQR} (\textbf{C}ontext-\textbf{a}ware \textbf{Q}uery \textbf{R}epresentation learning), that incorporates both \textit{structural context} and \textit{relation-induced context}. The structural context encodes the positional or role-like information of a node within the query computation graph. For example, in Figure~\ref{fig:figure1}, the structural context of node V could include the number of incoming and outgoing edges, the canonical position of V within the query graph, and the length of the query graph containing node V. Based on this, \textit{Figure} can be excluded as a V candidate because it is unlikely to be derived through the one-hop reasoning from the \textit{1980 Olympics} in KG. The relation-induced context, on the other hand, leverages KG entities linked to each node's relations. For instance, in the query graph shown in Figure~\ref{fig:figure1}, the relation-induced context—acquired by identifying nodes in the KG associated with $TeamsPlay^{-1}$—suggests that \textit{Sprint}, \textit{Marathon}, and \textit{Triathlon} are not suitable candidates for \textit{V} by accentuating the entities which linked to the $TeamsPlay^{-1}$ relations (\textit{Soccer, Volleyball,} and \textit{Basketball}). By incorporating structural and relation-induced contexts into the existing query embedding methods, our approach mitigates the cascading errors that can arise from the step-by-step computation characteristic of these methods.
Our contributions are threefold: 
(1) We propose a novel query representation learning method that leverages two types of context within a query computation graph: structural context and relation-induced context, which are often overlooked in previous methods. 
(2) Our proposed technique is applicable to any existing query embedding-based method, as it utilizes structural and relation-induced context acquired from the input query graph, irrespective of the models. 
(3) Our experiments show that our method leads to performance improvements for various models (Q2B~\cite{q2b}, BetaE~\cite{betae}, and ConE~\cite{cone}), which have received considerable attention in various FOL tasks as foundation models.
Specifically, our experiments on two benchmark datasets demonstrate that our method consistently improves these models, achieving up to a 19.5\% enhancement in query reasoning tasks compared to their baselines.

\section{Related Work}\label{related_work}


\paragraph{Geometry-based.} These approaches represent each query as a geometric shape in the embedding space. They then identify entities located close to that shape as answers to the query. For instance, GQE~\cite{gqe} maps entities to points and employs neural networks to model logical operators. Q2B~\cite{q2b} and its extension~\cite{newlook} use hyper-rectangles, or boxes, to represent queries. These boxes are defined by a center and offset, and entities positioned closer to them are considered potential answers. ConE~\cite{cone} represents queries using multiple two-dimensional cones characterized by their axis and aperture angle. 
Similar to Q2B, entities lying closer to the cone representing the query are considered potential answers. 
However, these geometry-based models fail to harness both structural and relation-induced context information.
\vspace{-2mm}
\paragraph{Probability-based.} In contrast to geometry-based approaches, probability-based methods represent queries as probability distributions. For example, BetaE~\cite{betae} uses multiple beta distributions to represent a query, entities, and relations. Similar approaches include GammaE~\cite{gammae} with gamma distributions and PERM~\cite{perm} with multivariate Gaussian distributions. These methods measure the distance between the query embedding and candidate entities using KL-divergence. 
FuzzQE~\cite{fuzzqe} takes a different approach, mapping queries and entities to a fuzzy space. It represents the answer set as a fuzzy vector and calculates the probability of an entity being an answer using a score function. 
Similarly, GNN-QE~\cite{gnnqe} adopts fuzzy logic to represent queries and entities but uses graph neural networks for projection in an incomplete KG.  On the other hand, WFRE~\cite{wfre} models queries and entities as discretized mass vectors that satisfy fuzzy logic. It models logical operators through t-norm or t-conorm functions and measures the distance between mass vectors using the optimal transport theory~\cite{optimal_transport}. While each method has its advantages and disadvantages and is selectively utilized in various FOL-based applications~\cite{logicrec,CBOX4CR,xiong2023geometric}, the potential of query-based context encoding to enhance multiple multi-hop reasoning approaches and yield robust performance improvements has not received sufficient attention.
\vspace{-2mm}
\paragraph{Query Encoder-based.} There are auxiliary transformation encoders for representing the FOL queries for multi-hop reasoning. These include  Q2T~\cite{q2t} and LMPNN~\cite{lmpnn}, which take distinct approaches to tackling complex queries. Q2T transforms the complex query into a single virtual triple of head, relation, and tail components. It computes the score of this virtual triple with a pre-trained KG embedding model to predict the answer to the query. LMPNN~\cite{lmpnn}, on the other hand, decomposes the complex multi-hop query into multiple simpler triples. It generates a message using the entity embedding to maximize the score for each triple, utilizing a pre-trained KG embedding model. 
However, these approaches require initial KG embeddings from extensive pre-training and are not applicable to various multi-hop reasoning approaches.

\section{Preliminaries} \label{preliminaries}

Given an entity set $\mathcal{V}$ and a relation set $\mathcal{R}$, the KG $G = {(h, r, t)} \subset \mathcal{V} \times \mathcal{R} \times \mathcal{V}$ represents a collection of triples that encapsulate factual information in the real world. Here, $h, t \in \mathcal{V}$, and $r \in \mathcal{R}$. When considering each relation as a binary function, such as $r(h, t)$, following the structure of predicate logic, the triples observed from the KG hold a value of \textit{True}.

\subsection{First-Order Logic (FOL) Queries}
The First-Order Logic (FOL) queries take predicate logic, allowing the use of quantified variables. 
A FOL query is composed of a non-variable anchor entity set $\mathcal{V}_{a} \subseteq \mathcal{V}$, an existential quantified set $\{V_1, ... V_k\}$ of size $k$, and a target variable $V_?$, which is an answer to a certain query. 
For example, in Figure~\ref{fig:figure1}, $\mathcal{V}_{a}$ is $\{\text{1980 Olmypic}\}$, the existential quantified set is $\{V\}$, and the target variable is $A$. Generally, FOL queries include four logical operators: the existential quantifier($\exists$), conjunction($\wedge$), disjunction($\vee$), and negation($\neg$). We can formulate a FOL query $q$ as follows: 


\vspace{-2mm}
\begin{eqnarray}\label{eq1}
\let\frac\dispfrac
\let\theHequation\theequation
\label{dfg-b03adfec9076}
\begin{array}{@{}l}\begin{array}{l}q=V_?.\;\exists V_1,...,V_k:\;c_1\vee c_2\vee...\vee c_n.\\\end{array}\end{array}
\end{eqnarray}
\vspace{-6mm}
\begin{align}
    c_i = a_{i_1} \wedge a_{i_2} \wedge ... \wedge a_{i_m}.
\end{align}
Each atomic formula $a_{i_j}$ is in the form of predicate logic which has the form of $r(v_a, V)$ or $\neg r(v_a, V)$ or $r(V', V)$ or $\neg r(V', V)$ consisting a conjunction $c_i$ of $m$ predicates. $v_a$ is an element of $\mathcal{V}_a$. $V$ and $V'$ are elements of $\{V_?, V_1, ..., V_k\}$ and $\{V_1, ..., V_k\}$, respectively ($V' \neq V$). Note that we call this atomic formula $a_{i_j}$ a \textit{branch} of a query graph.



\subsection{Query Graph}
In BetaE~\cite{betae}, FOL query answering tasks are categorized into pre-defined multi-hop reasoning tasks using 14 different query types.
Each query type can be represented by a corresponding query graph and all queries can be interpreted as corresponding natural language questions. 
For instance, a FOL query for \textit{``What are the names of the teams playing the team sports featured in the 1980 Olympics?''} can be represented by the query graph shown in Figure~\ref{fig:figure1}. 
A query graph consists of anchor nodes, variable nodes, and answer nodes. Variable nodes signify entities fulfilling individual sub-conditions within the query, while the answer node represents the entity satisfying the entire query.
In Figure~\ref{qtype_part}, you can observe five types of query graphs, and all query types can be found in Appendix~\ref{query_dataset}. 


\begin{figure}[t]
    \centering
    \includegraphics[width=0.8\columnwidth]{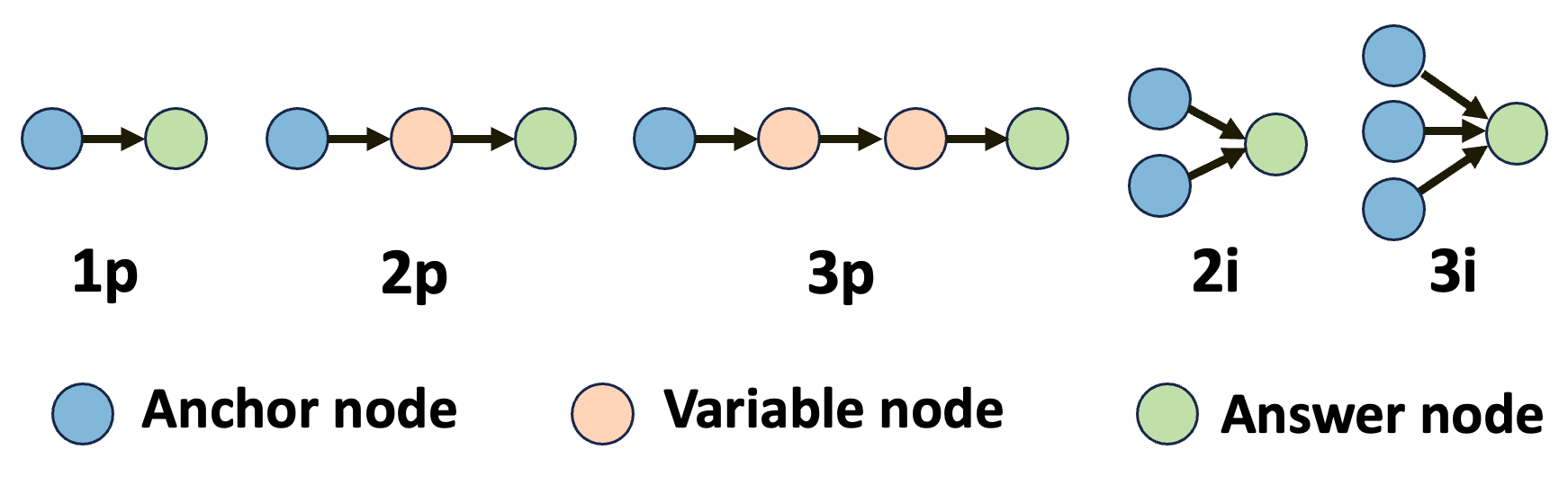}
    \vspace{-2mm}
    \caption{Five types of query graph}
    \label{qtype_part}
    \vspace{-2mm}
\end{figure}


\subsection{Computation Graph} 


The computation graph details the computation procedure to obtain the embedding of each node in the query graph. In a computation graph, each node represents an embedding of an entity (or a set of entities) in the KG, and each edge signifies a logical transformation (e.g., relational projection, intersection/union/negation operators) of this distribution. The computation graph for a FOL query resembles a tree. The root node of the computation graph represents the answer (or target) variable, with one or more anchor nodes provided by the FOL. Embeddings of entities and transformation operators are initialized; embeddings of anchor nodes are then fed into the neural network of logical operators in a serial manner to obtain the final embeddings for the answer variable, thereby creating a query embedding. During training, models ensure the proximity of query embeddings to the ground truth. In the prediction stage, entities close to the query embedding are utilized for prediction. Further details on the logical operators are included in the Appendix~\ref{computation_graph}.

\section{Methodology}\label{methodology}

\begin{figure*}[ht]
\centering
\includegraphics[width=\textwidth]{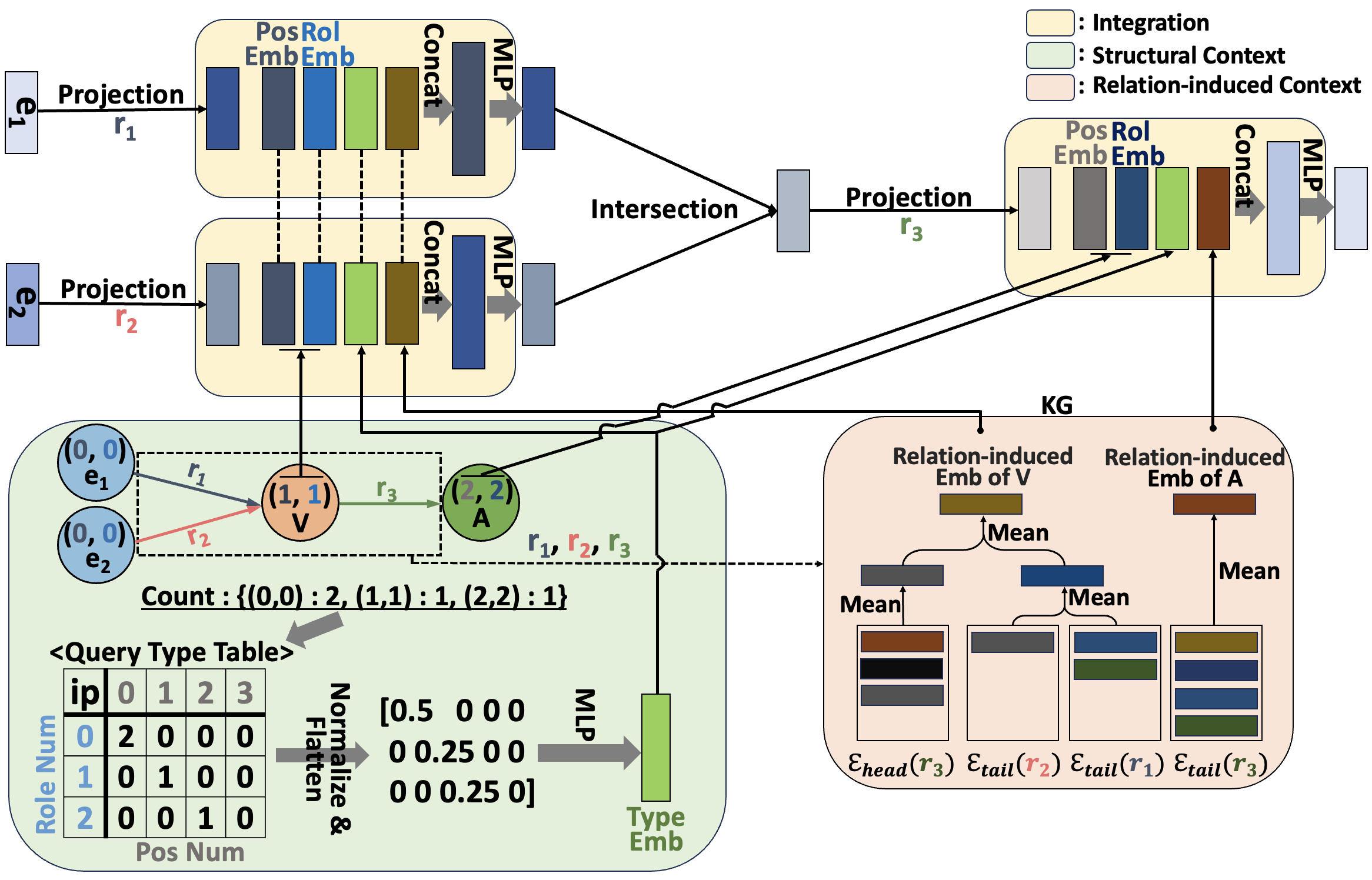}  
\vspace{-4mm}
\label{fig:meta_semantic_info}
\caption{The figure of the structural context and relation-induced context and its application example on \textit{ip} query. 
Each node in the query graph can be assigned a position number and a role number, which is represented as a tuple in the \textit{Structural Context box (green box)}.
The first number of the tuple in each node of the query graph represents the \textit{position} number, and the second number indicates the \textit{role} number. Type embedding is derived from the query-type table containing the position and role information of the corresponding query graph. The \textit{Relation-induced Context} box illustrates constructing relation-induced embedding of node $V$ and $A$ from KG. The \textit{Integration} box describes integrating query embedding, position embedding, role embedding, type embedding, and relation-induced embedding into updated query embedding. Best viewed in color.}
\label{figure2}
\vspace{-4mm}
\end{figure*}
In this section, we explain \mname{}, our method that leverages context information from the query graph to create more accurate and nuanced entity embeddings. By incorporating context, \mname{} can make fine-grained adjustments within the embedding space, capturing the complexities of real-world queries. 

\mname{} concentrates on two key types of context: \textit{structural context} and \textit{relation-induced context}. The structural context combines positional cues (where nodes appear in the query) and functional roles (the specific role a node plays) to comprehend the overall structure and relationships within the query. Furthermore, the relation-induced context focuses on the specific relationships and interactions between entities within the query. \mname{} combines these structural and relation-induced contexts to create a more comprehensive query embedding.

\subsection{Learning Structural Context}

To capture structural context, we propose using position embeddings, role embeddings, and type embeddings for differentiating between different query types.
Note that position, role, and type embeddings are generated even for query types not in training data. 


\subsubsection{Position Embedding}
Previous studies \cite{lspe, pgnn} have shown the advantages of incorporating canonical positioning information within message-passing frameworks to present nodes in arbitrary graphs. Inspired by this, we propose incorporating the canonical positioning information in the representation of each node within the query graph, aiming to lead to a more expressive and informative representation. 

To capture the relative order of nodes within a query graph, we introduce the concept of canonical node positions. These positions represent the order in which nodes would appear if the graph were listed from left to right, as illustrated in Figure~\ref{figure2} (the first number in each tuple). Importantly, the maximum canonical position number remains consistent across all queries of the same type and equals the maximum query length. For example, consider the \textit{3p} query graph in Figure~\ref{qtype_part}. When listing nodes from left to right (starting with the anchor node), the maximum position is 4. Therefore, we initialize position embeddings for canonical positions 0 to 3 as follows: 


\begin{eqnarray}
\mathbf{P}_{pos} = \begin{bmatrix}\mathbf{p_0} & \mathbf{p_1} & \mathbf{p_2} & \mathbf{p_3} \end{bmatrix} \in \mathbb{R}^{4 \times d_{pos}},
\end{eqnarray}

\noindent where $d_{pos}$ denotes the dimension of the position embedding, and $\mathbf{p}_i$ represents the position embedding of a node at the $i$-th canonical position on the query graph. The lookup table $\mathbf{P}_{pos}$ maps each potential canonical position $i$ to its corresponding embedding vector $\mathbf{p}_i$. Then, $\mathbf{p}_i$ is integrated into the embedding of the node located at the $i$-th position on the query graph. The integration of this position embedding aims to compensate for the lack of explicit structural information arising from the sequential nature of existing embedding models.


\subsubsection{Role Embedding}

Query graphs reveal important structural information through distinct node roles. 
Regardless of the type of query, anchor nodes never have incoming edges, as shown in Figure~\ref{qtype_part}. In contrast, variable nodes are connected to other nodes from both incoming and outgoing edges. Finally, answer nodes serve as leaf nodes, with no outgoing edges. 
These distinct node roles, observed across various query graph types, offer valuable insights into the query's structure, enriching the information available for the embedding model.


To take advantage of this information, we assign a canonical number to each role of the query graph (i.e., 0 for the anchor node, 1 for the variable node, 2 for the answer node) and initialize the role embedding for each role as: 
\begin{eqnarray}
    \mathbf{R}_{rol} = \begin{bmatrix}\mathbf{r}_{anch} & \mathbf{r}_{var} & \mathbf{r}_{ans} \end{bmatrix} \in \mathbb{R}^{3 \times d_{rol}},
\end{eqnarray}
\noindent where $d_{rol}$ represents the dimension of the role embedding. $\mathbf{R}_{rol}$ is a lookup table containing role embeddings. $\mathbf{r}_{anch}$, $\mathbf{r}_{var}$, and $\mathbf{r}_{ans}$ are role embeddings for anchor, variable, and answer nodes, respectively. The role embedding is incorporated into the query embeddings of the corresponding node to provide information about the node's role within the query graph.



\subsubsection{Type Embedding}

While position and role embeddings provide valuable information about each node within a query graph, they don't necessarily reveal the complete picture of how these nodes interact and contribute to the distinct query structures. For instance, they may struggle to represent unique query types like \textit{2i} and \textit{3i} in Figure~\ref{qtype_part}. The answer entities in both \textit{2i} query and \textit{3i} query have position 1 with respect to the anchor entity (\textit{position 0}) and have the same role embedding (\textit{answer}). Therefore, position embedding and role embedding are not enough to reflect the different structure information of \textit{2i} and \textit{3i} query graphs.

To address this limitation, we propose a method to capture the structural context of a given query type. This approach extends beyond individual node properties to understand the broader relationships and dependencies between them.
To capture this crucial structural context, we introduce the query-type table. Illustrated in the \textit{Structural Context} box of Figure~\ref{figure2}, this table encodes the complete structural information of the query by representing each node as a combination of its position and role. Flattening this table results in a vector, where each element corresponds to a specific position-role combination. Consequently, this vector uniquely identifies the query's structure (i.e., query type).
Because the range of values in the query type table is different for each query type, we normalize the vector by dividing by 4, which is the maximum sum of the values in the vector.
Finally, we use a linear transformation to map this normalized vector into a continuous embedding space, creating the type embedding. For a query graph $\mathcal{G}$, the type embedding of any node in the query graph $\mathcal{G}$ is presented as follows:


\begin{equation}
    \mathbf{g}_{\mathcal{G}} = \mathbf{W}_g \cdot flatten(Count(\mathcal{G})).
\end{equation}$Count$ returns a query-type table ($Count(\mathcal{G}) \in \mathbb{R}^{3 \times 4}$), constructed by counting the occurrence of each combination of position number and role number for the nodes in the graph $\mathcal{G}$. The $flatten$ function transforms this two-dimensional query-type table into a vector. $\mathbf{W}_g$ represents a linear transformation matrix. The type embedding is then incorporated into the query embedding of each node within a query graph. All nodes within a query graph share the same query table, resulting in identical type embedding. It's worth noting that we collectively refer to the position, role, and type embeddings as \textit{structure embeddings}.

\subsection{Learning Relation-induced Context}

Several approaches have been proposed for leveraging neighboring relations around entities to enhance reasoning in KGs~\cite{faan, morse}.
However, computing attention scores, as in ~\cite{faan}, requires a high computational cost, making it more expensive to apply these approaches concurrently with computing the query embedding.
Additionally, it is challenging to represent an entity solely based on relations' embedding as in ~\cite{morse} due to the distinct distributions of entities and relations. 
To extract relation-induced context associated with the nodes, we first define $\mathcal{N}_{r}^{in}(v_i)$ as the set of incoming relations of the node $v_i$ in the query graph. $\mathcal{N}_{r}^{out}(v_i)$ represents the set of outgoing relations of the node $v_i$ in the query graph. We then utilize the connection of the relation in the KG to acquire the information possessed by the relation associated with node $v_i$ on the query graph. 
To achieve this, we define the entities serving as the tail of relation $r$ in the KG as $\mathcal{E}_{tail}(r)$, and the entities serving as the head as $\mathcal{E}_{head}(r)$. We then aggregate the embeddings of these entities to construct the relation-induced context of the node, referred to as the relation-induced embedding. This can be formulated as follows:

\vspace{-2mm}
\begin{equation}\label{lvin}
    \mathbf{l}_{v}^{in}=Agg\{Emb(e)\,|\,e \in \mathcal{E}_{tail}(r),\,r \in \mathcal{N}_{r}^{in}(v)\},
\end{equation}
\begin{equation}\label{lvout}
    \mathbf{l}_{v}^{out}=Agg\{Emb(e)\,|\,e \in \mathcal{E}_{head}(r),\,r \in \mathcal{N}_{r}^{out}(v)\},
\end{equation}
\begin{equation}
    \mathbf{l}_v=(\mathbf{l}_{v}^{in} + \mathbf{l}_{v}^{out})/2,
\end{equation}where we use the mean operation as the $Agg$ function, and $\mathbf{l}_v$ indicates the relation-induced embedding of node $v$. Note that in Equation~(\ref{lvin}) and (\ref{lvout}), considering computational efficiency, we sample up to $K$ number of entities from $\mathcal{E}_{tail}(r)$ and $\mathcal{E}_{head}(r)$, respectively. 
\subsection{Integrating Two Contextual Embeddings}\label{integration}
We incorporate the structural embedding (Sec 4.1) and the relation-induced embedding (Sec 4.2) into the embeddings of each node in the query graph using a neural network architecture. Different embeddings are passed through the neural network, concatenated, and then mapped into the same low-dimensional embedding space as the original query embedding. The node embedding of the arbitrary query graph $\mathcal{G}$, which incorporates the obtained information from above, can be formulated as follows:
\begin{equation}
    \mathbf{p}_{v} = \mathbf{P}_{pos}[position_v],
\end{equation}
\begin{equation}
    \mathbf{r}_{v} = \mathbf{R}_{rol}[role_v], 
\end{equation}
\begin{equation}\label{integration1}
    \mathbf{q}_{v}' = \mathbf{W'}\cdot (\mathbf{MLP}_q \cdot \mathbf{q}_{v} \, | \, \mathbf{MLP}_I(\{\mathbf{p}_v, \mathbf{r}_v, \mathbf{g}_{\mathcal{G}}, \mathbf{l}_v\}),
\end{equation}
where $[ \cdot ]$ is a lookup operation. 
Note that the node $v$ can be either a variable node or an answer node in $\mathcal{G}$. $position_v$ denotes the canonical position of node $v$ in the query graph $\mathcal{G}$. $role_v$ represents the number corresponding to the role of node $v$ in the query graph $\mathcal{G}$.
$\mathbf{p}_v$, $\mathbf{r}_v$, $\mathbf{g}_{\mathcal{G}}$, and $\mathbf{l}_v$ represent the position embedding, role embedding, type embedding, and the relation-induced embedding of node $v$, respectively. $\mathbf{MLP}_q$ and $\mathbf{MLP}_I$ signify multi-layer perceptrons, respectively.
$(\cdot|\cdot)$ is a concatenation. 
$\mathbf{q}_v$ represents the embedding of the branch toward node $v$ in the query graph resulting from projection operators of the base query embedding model (e.g., Q2B, BetaE, or ConE). $\mathbf{q}_v'$ represents the updated representation obtained by integrating structural and relation-induced context into $\mathbf{q}_v$.

The query embedding models sequentially compute embeddings of nodes in the query graph, ultimately deriving the embedding for the answer node. The integration occurs at each instance of a projection operation. The position, role, type, and relation-induced embeddings are precomputed for each variable node and answer node. During the execution of projection operation, these context embeddings are integrated into the embedding of the corresponding node in the query graph, resulting in an enhanced representation.

\subsection{Training}\label{training}
The structure embeddings, relation-induced embedding and parameters consisting of neural networks of Equation~(\ref{integration1}) for integration are updated through backward propagation from the loss function of each query embedding model.
The loss function defined in the query embedding model is common as follows:
\begin{align}\label{loss}
\begin{split}
L_{qe} = & -\log \sigma(\gamma - \text{Dist}(\mathbf{v}, \mathbf{q}_A')) \\
& - \sum_{j=1}^{k} \frac{1}{k} \log \sigma(\text{Dist}(\mathbf{v_j'}, \mathbf{q}_A')-\gamma).
\end{split}
\end{align}Here, $\mathbf{q}_A'$ indicates the modified query embedding of the answer node by Equation~(\ref{integration1}). $\mathbf{v}$ and $\mathbf{v_j'}$ denote the positive (i.e., answer entity) and negative entity for each query. $k$ represents the number of negative samples~\cite {word2vec}, and $Dist$ is the model-specific function that measures the distance between query embedding and entity embedding. \\
In the case of BetaE, utilizing entities' embeddings as parameters for beta distributions can lead to significant variance shifts, hindering the convergence of the model's learning. Therefore, for BetaE, we introduce an additional loss to mitigate the variance shifts. The details about the variance loss are provided in Appendix~\ref{appendix_variance}.

\subsection{Time Complexity Analysis}
Acquiring the relation-induced embedding costs $\mathcal{O}(NKD)$, where $N$ is the number of nodes in the query graph, $K$ is the size of sample entities for constructing the relation-induced embedding, and $D$ is the dimension of the embedding.
When integrating all embeddings, $\mathcal{O}(PMD^2 + NKD) \approx \mathcal{O}({1 \cdot D^2})$ is required. Here, $P$ is the number of projection operations in the given query, and $M$ is the number of contextual information embeddings used (4 in case when using all embeddings). In addition to the analysis, we compare the inference times of the query embedding model and that with our methodology in Table~\ref{inference_time} of Appendix.

\section{Experiments}\label{experiment}
\begin{table*}[t]
\centering

\begin{adjustbox}{width=\linewidth,center}

{
\begin{tabular}{ccccccccccc|c|c}

\hline
\textbf{Dataset} & \textbf{Model} & \textbf{1p} & \textbf{2p} & \textbf{3p} & \textbf{2i} & \textbf{3i} & \textbf{pi} & \textbf{ip} & \textbf{2u} & \textbf{up} & \textbf{$Avg$} & \textbf{$Imp (\%) $}\\

\hline
\hline

\multirow{12}{*}{FB15k-237} & \multirow{4}{*} {} 
Q2B & 40.35 & 9.27 & 6.87 & 28.57 & 40.88 & 20.71 & 12.75 & 10.96 & 7.48 & 19.76 & - \\
\cline{2-13}
& Q2B+\strc{} & 42.49 & 10.82 & 8.92 & 32.55 & 45.92 & 23.44 & 13.29 & 14.27 & 8.67 & 22.26 & 12.7 \\
& Q2B+\local{} & 42.66 & 10.94 & 9.18 & 32.53 & 45.98 & 22.00 & 13.47 & 14.46 & 8.45 & 22.18 & 12.2 \\
& Q2B+\both & 42.65 & 10.82 & 9.00 & 32.43 & 46.42 & 21.70 & 13.39 & 14.53 & 8.75 & 22.19 & 12.3 \\
\cline{2-13}

& \multirow{4}{*} {} 
BetaE & 39.20 & 10.69 & 10.15 & 29.16 & 42.59 & 22.53 & 12.28 & 12.52 & 9.73 & 20.98 & - \\
\cline{2-13}
& BetaE+\strc{} & 40.82 & 12.16 & 10.78 & 32.04 & 45.80 & 24.66 & 13.71 & 13.94 & 10.59 & 22.72 & 8.3 \\
& BetaE+\local{} & 41.08 & 12.31 & 10.70 & 31.64 & 45.21 & 24.49 & 14.22 & 13.94 & 10.46 & 22.67 & 8.1 \\
& BetaE+\both{} & 41.35 & 12.75 & 10.79 & 31.89 & 45.55 & 24.70 & \textbf{14.69} & 14.24 & 10.67 & 22.96 & 9.4 \\
\cline{2-13}

& \multirow{4}{*} {} 
ConE & 42.33 & 12.78 & 10.97 & 32.59 & 47.13 & 25.45 & 13.71 & 14.35 & 10.50 & 23.31 & - \\
\cline{2-13}
& ConE+\strc{} & 42.89 & \textbf{13.13} & \textbf{11.16} & 33.07 & 47.56 & 24.75 & 11.97 & \textbf{15.49} & 10.90 & 23.44 & 0.5 \\
& ConE+\local{} & {\textbf{42.92}} & 12.98 & \textbf{11.16} & 32.32 & 46.99 & 24.31 & 14.36 & 15.18 & 10.67 & 23.43 & 0.5 \\
& ConE+\both{} & 41.81 & 12.84 & 10.90 & \textbf{33.78} & \textbf{48.24} & \textbf{25.65} & 14.41 & 14.48 & \textbf{10.95} & \textbf{23.67} & 1.5 \\

\hline
\hline
\multirow{12}{*}{NELL} & \multirow{4}{*} {} 
Q2B & 41.48 & 13.80 & 11.17 & 32.01 & 44.71 & 21.90 & 16.80 & 11.19 & 10.12 & 22.57 & - \\ 
\cline{2-13}
& Q2B+\strc{} & 57.12 & 16.15 & 13.54 & 37.81 & 50.92 & 23.45 & 17.47 & 14.60 & 10.37 & 26.83 & 18.9 \\
& Q2B+\local{} & \textbf{57.15} & 16.08 & 13.75 & 37.84 & 50.68 & 22.82 & \textbf{18.03} & 14.80 & 10.36 & 26.83 & 18.9 \\
& Q2B+\both{} & \textbf{57.15} & 15.99 & 13.42 & 37.95 & \textbf{51.25} & 24.24 & 17.57 & 14.80 & 10.42 & 26.97 & 19.5 \\
\cline{2-13}

& \multirow{4}{*} {} 
BetaE & 53.09 & 13.10 & 11.56 & 37.62 & 47.87 & 24.40 & 14.89 & 12.05 & 8.63 & 24.80 & - \\
\cline{2-13}
& BetaE+\strc{} & 54.72 & 14.83 & 12.58 & 37.49 & 48.34 & 24.06 & 15.00 & 12.50 & 9.76 & 25.47 & 2.7 \\
& BetaE+\local{} & 54.42 & 14.44 & 12.53 & 37.03 & 47.76 & 24.01 & 15.64 & 12.08 & 9.48 & 25.26 & 1.9 \\
& BetaE+\both{} & 55.20 & 15.44 & 13.52 & 37.99 & 48.47 & 25.45 & 16.59 & 13.06 & 10.26 & 26.22 & 5.7 \\
\cline{2-13}

& \multirow{4}{*} {} 
ConE & 53.19 & 16.08 & 14.04 & 39.88 & 51.08 & \textbf{26.04} & 17.58 & \textbf{15.41} & 11.24 & 27.17 & - \\
\cline{2-13}
& ConE+\strc{} & 55.86 & 17.27 & \textbf{14.95} & 39.84 & 50.91 & 24.77 & 17.76 & 14.86 & 11.95 & \textbf{27.58} & 1.5 \\
& ConE+\local{} & 55.28 & 16.94 & 14.77 & 39.95 & 50.98 & 24.31 & 16.42 & 15.02 & 11.41 & 27.23 & 0.2 \\
& ConE+\both{} & 56.05 & \textbf{17.49} & 14.57 & \textbf{40.46} & 51.08 & 22.99 & 16.50 & 15.06 & \textbf{11.99} & 27.35 & 0.7\\


\hline

\end{tabular}
}

\end{adjustbox}

\hrule height 0pt
\vspace{2mm}
\caption{MRR results (\%) for answering conjunctive queries without negation ($\exists$, $\wedge$, $\vee$) on FB15k-237 and NELL. $Imp$ denotes the percentage of improvement in average MRR compared to the base reasoning models. +\strc{} denotes the model utilizing structure embedding. +\local{} indicates the model using relation-induced embedding. +\both{} represents the model using both structure and relation-induced embeddings. The bold text highlights the best result for each type of query.} 
\label{main_results}
\vspace{-4mm}
\end{table*}

\begin{figure*}[t]
    \centering
    \begin{subfigure}{0.32\linewidth} 
        \centering
        \includegraphics[width=\linewidth]{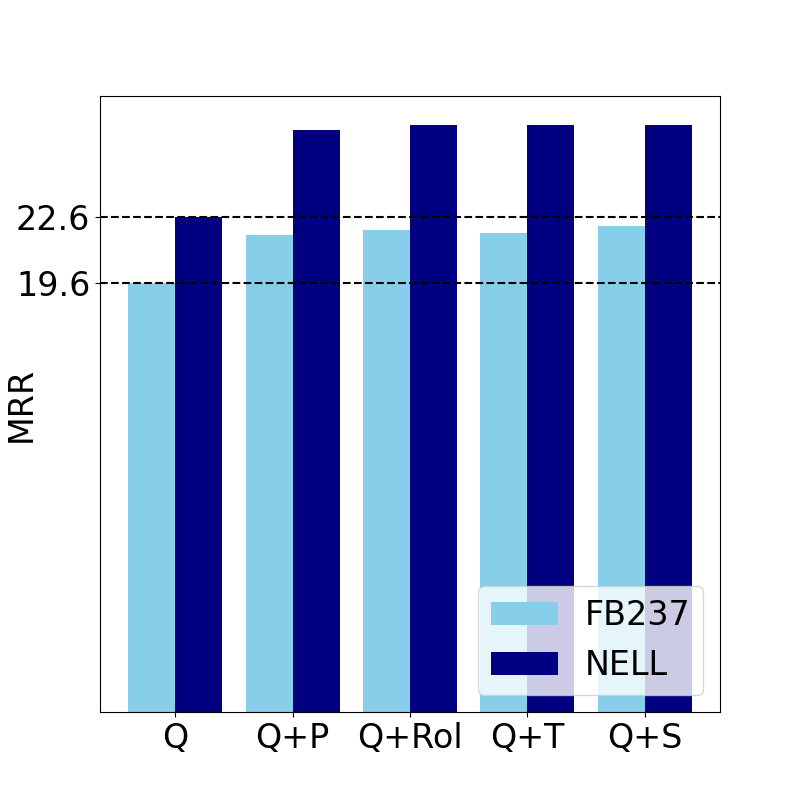}
        \caption{Q2B}
        \label{abla_q2b}
    \end{subfigure}\hfill 
    \begin{subfigure}{0.32\linewidth} 
        \centering
        \includegraphics[width=\linewidth]{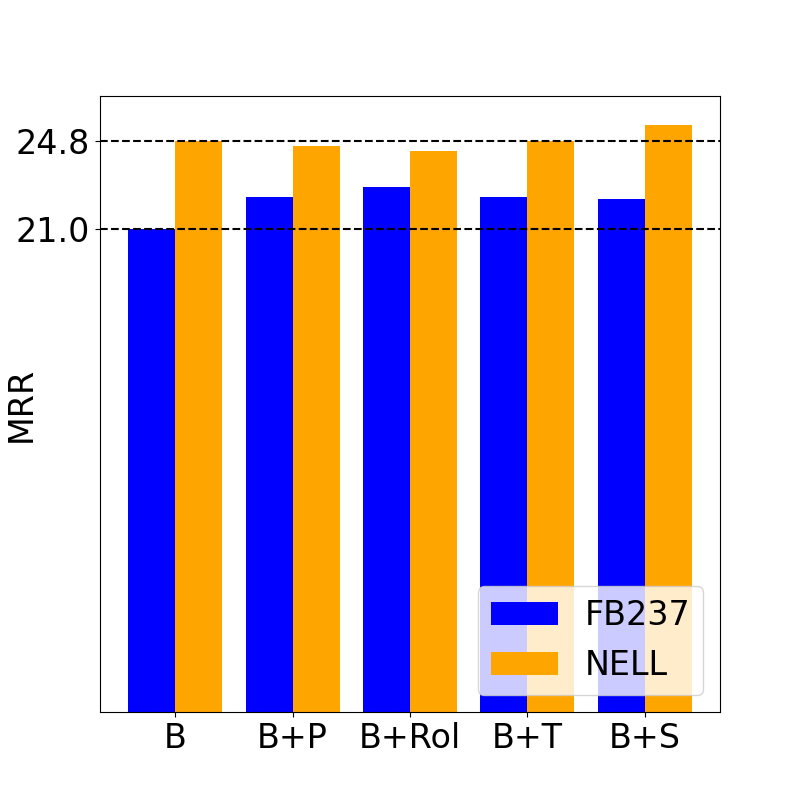}
        \caption{BetaE}
        \label{abla_betae}
    \end{subfigure}
    \begin{subfigure}{0.32\linewidth}
        \centering
        \includegraphics[width=\linewidth]{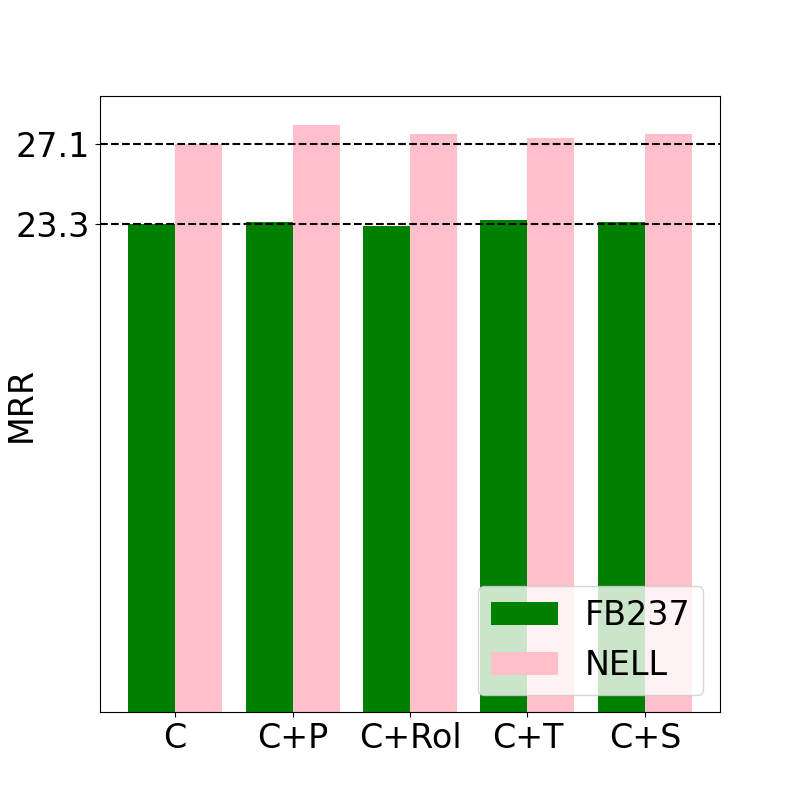}
        \caption{ConE}
        \label{abla_cone}
    \end{subfigure}
    \vspace{-2mm}
    \caption{Ablation study on the existence of Position, Role, and Type embedding. FB237 denotes the FB15k-237 dataset. +P, +Rol, +T and +S indicates the model with position embedding, role embedding, type embedding, and all the structure embeddings, respectively.}
    \label{ablation}
    \vspace{-4mm}
\end{figure*}

{
\begin{table}[t]
\centering
\resizebox{\columnwidth}{!}{

\begin{tabular}{ccccccc|c}
\hline

\textbf{Dataset} & \textbf{Model} & \textbf{2in} & \textbf{3in} & \textbf{inp} & \textbf{pni} & \textbf{pin} & \textbf{$Avg$}\\

\hline
\hline

\multirow{8}{*}{FB15k-237} & \multirow{4}{*} {} 
BetaE & 5.19 & 7.94 & 7.44 & 3.60 & 3.55 & 5.54 \\
\cline{2-8}
& BetaE+\strc{} & 5.29 & 8.35 & 7.77 & 3.51 & 3.90 & 5.76 \\
& BetaE+\local{} & 5.42 & 8.30 & 7.83 & 3.60 & 3.89 & 5.81 \\
& BetaE+\both{} & 5.49 & 8.24 & 8.01 & 3.77 & 3.91 & 5.88\\
\cline{2-8}

& \multirow{4}{*} {} 
ConE & 5.78 & 9.57 & 7.95 & 3.85 & 4.41 & 6.31 \\
\cline{2-8}
& ConE+\strc{} & 6.25 & 9.62 & 7.60 & 3.97 & 4.03 & 6.30 \\
& ConE+\local{} & \textbf{6.84} & \textbf{10.44} & 7.66 & \textbf{4.49} & \textbf{4.79} & \textbf{6.85} \\
& ConE+\both{} & 5.24 & 8.91 & \textbf{8.07} & 3.51 & 3.90 & 5.93 \\

\hline
\hline
\multirow{8}{*}{NELL} & \multirow{4}{*} {} 

BetaE & 5.23 & 7.64 & 10.12 & 3.27 & 3.09 & 5.87 \\
\cline{2-8}
& BetaE+\strc{} &  5.37 & 7.96 & 10.43 & 3.46 & 3.29 & 6.10 \\
& BetaE+\local{} & 4.91 & 7.63 & 10.06 & 3.17 & 3.02 & 5.76 \\
& BetaE+\both{} & 5.50 & 7.62 & 10.87 & 3.70 & 3.30 & 6.20  \\
\cline{2-8}

& \multirow{4}{*} {} 
ConE & 5.70 & 8.01 & 10.96 & 3.83 & 3.58 & 6.40  \\
\cline{2-8}
& ConE+\strc{} & 6.02 & 8.11 & \textbf{11.54} & 4.01 & \textbf{3.72} & \textbf{6.88} \\
& ConE+\local{} & \textbf{6.10} & \textbf{8.35} & 11.02 & \textbf{4.07} & 3.65 & 6.64 \\
& ConE+\both{} & 5.77 & 8.01 & 11.12 & 3.87 & 3.66 & 6.49 \\

\hline
\end{tabular}}
\caption{MRR results (\%) for answering conjunctive queries with negation.} 
\vspace{-4mm}
\label{negation_results}
\end{table}
}


\subsection{Datasets and Base Reasoning Models} 

For evaluation, we use two datasets: FB15k-237~\cite{fb15k-237}  and NELL~\cite{nell995}.
We evaluate the effectiveness of our approach on the Q2B~\cite{q2b}, BetaE~\cite{betae}, and ConE~\cite{cone} models, which have received considerable attention in various FOL tasks as foundation models. For more detailed information on the datasets and query structures, please refer to Appendix~\ref{appendix}.



\subsection{Main Results}

Table~\ref{main_results} demonstrates that applying our method results in performance gains for all compared models, with Q2B showing a particularly notable 19.5\% increase on the NELL dataset. This significant improvement could stem from the unique alignment between Q2B's hyper-rectangular embedding space and the space where our position, role, and type embeddings are mapped.


The lower performance of BetaE+\local{} compared to BetaE+\strc{} can be attributed to the parameters of BetaE. BetaE utilizes two d-dimensional embedding vectors, \boldsymbol{$\alpha$} and \boldsymbol{$\beta$}, as the parameters of beta distributions, to represent entities. The relation-induced embedding is constructed for each query node by computing the average values across multiple entities. However, since the parameter of averaged multiple beta distributions is not the same as the average of parameters of beta distributions, the resulting relation-induced embedding can deviate from the actual values.

While not as pronounced as with Q2B and BetaE, our method still delivers performance gains for ConE across 1p, 2p, and 3p queries. This suggests that our methodology benefits even simpler query structures, indicating its broad applicability and effectiveness.
Table~\ref{main_results} demonstrates a consistent performance improvement with the application of \strc{}. Furthermore, in the context of complex queries (e.g., \textit{2i}, \textit{3i}, \textit{pi}, and \textit{ip}), a synergistic effect is evident when both S and R are employed.

Table~\ref{negation_results} presents the experimental results for queries with negation. While there is an overall positive effect, it is smaller compared to previous query types. This could be attributed to the late application of our method, which operates after the projection step.



\subsection{Ablation Study}\label{ablation_study}
We conduct an ablation experiment on Q2B, BetaE, and ConE to dissect the roles of position, role, and type embeddings in building the structural context of the query graph, as shown in Figure~\ref{ablation}. Interestingly, the results reveal that using any single embedding (position, role, or type) achieves performance comparable to using the combination of all three, demonstrating the effectiveness of each embedding option.


\subsection{Hyper-parameter Sensitivity Study}
\begin{figure}[t]
    \centering
    \begin{subfigure}{0.5\columnwidth} 
        \centering
        \includegraphics[width=\linewidth]{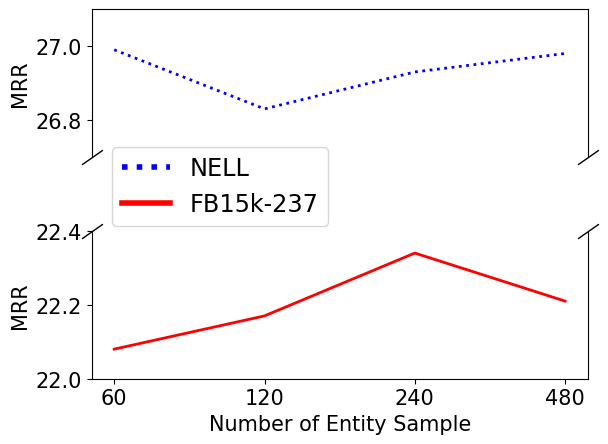}
        \caption{Q2B+\local{}}
        \label{sample}
    \end{subfigure}\hfill 
    \begin{subfigure}{0.5\columnwidth} 
        \centering
        \includegraphics[width=\linewidth]{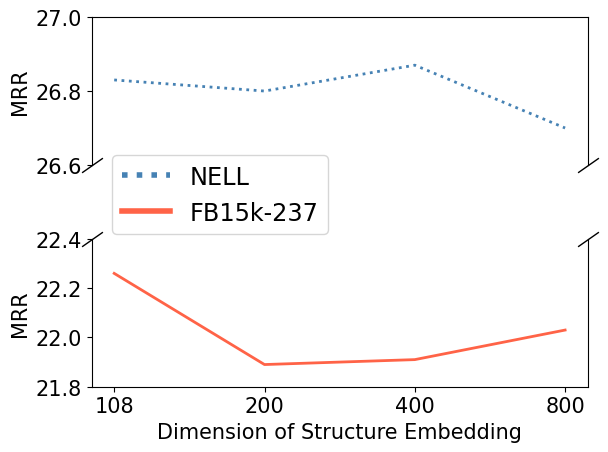}
        \caption{Q2B+\strc{}}
        \label{dim}
    \end{subfigure}
    
    \vspace{-2mm}
    \caption{Effect of Hyper-parameters on Q2B+\{\strc{}, \local{}\}.}
    \label{hypesense}
    \vspace{-4mm}
\end{figure}

Two considerations arise when applying our methodology, \mname{}, to a query embedding-based model: the number of entities sampled to construct the relation-induced embedding and the dimension of the structure embedding.
\paragraph{Size of Entity Samples.}\label{sample_entity}
We evaluate the effect of entity samples per relation when constructing relation-induced embedding from KG on Q2B+\local{}. The number of entity samples varies to 60, 120, 240, and 480. The results are depicted in Figure~\ref{hypesense} (a). For FB15k-237, we observe an improvement in performance on Q2B as the sample size increases, along with a notable variation in performance based on sample size. However, in the case of NELL, it is evident that the model's performance is relatively insensitive to the number of samples. The number of entities on the NELL dataset could be the factor of this result because the FB15k-237 dataset has a smaller number of entities than the NELL dataset.

\paragraph{Dimension of Structure Embeddings.}

Furthermore, we conduct a hyper-parameter sensitivity study on Q2B+\strc{}, specifically focusing on the effect of position embedding, role embedding, and type embedding (i.e., structure embedding) dimensions. The dimensions are varied to 108, 200, 400, and 800, respectively. The results are presented in Figure~\ref{hypesense} (b). For FB15k-237, the most optimal performance was observed with a dimension of 108 while significant performance degradation was evident with other dimensions. In the case of NELL, it is evident that the model's performance is relatively insensitive to changes in dimension.

\section{Conclusion}

In this paper, we introduced \mname{}, a model-agnostic approach that leverages structural and relation-induced context within the query graph. By seamlessly fusing position, role, type, and relation-induced embeddings into the query representation, \mname{} substantially improves the performance of query embedding-based multi-hop reasoning models across diverse query types.

\section{Ethics Statement}
Our model proposes a methodology that enhances the effectiveness of multi-hop logical reasoning by fully integrating the context of the FOL query graph. We do not additionally leverage any external knowledge or information that might bias the evaluation of our model. However, while this approach can improve model performance, it may also inadvertently reinforce existing harmful biases in the knowledge graphs.
\section{Limitation}
With the two context modeling approaches, we have observed an increase in performance. Although these approaches are model-agnostic, they may offer relatively smaller improvements, similar to those seen in ConE. Moreover, they introduce additional hyperparameters, such as the dimensions of structure and relation-induced embeddings, which can result in 1-2\% fluctuation in Mean Reciprocal Rank (MRR).
\section{Acknowledgement}
This work was supported by the Institute of Information \& Communications Technology Planning \& Evaluation (IITP) grant funded by the Korea government (MSIT): (No. 2019-0-00421, Artificial Intelligence Graduate School Program (Sungkyunkwan University)) and (No. RS-2023-00225441, Knowledge Information Structure Technology for the Multiple Variations of Digital Assets). This research was also supported by the Culture, Sports, and Tourism R\&D Program through the Korea Creative Content Agency grant funded by the Ministry of Culture, Sports and Tourism in 2024 (Project Name: Research on neural watermark technology for copyright protection of generative AI 3D content, RS-2024-00348469, 25\%)

\newpage
\bibliography{main}

\clearpage

\appendix
\section{Experimental Details}\label{appendix}

We use the default parameters from the existing baseline. The hyperparameter search was conducted for the number of entity samples \{60, 120, 240, 480\} and for the dimension of the structure embedding \{108, 200, 400, 800\}. We use 120 for the entity sample size, 800 for structure embedding for ConE~\cite{cone}, and 108 for the other baselines.

\subsection{Datasets} We use two datasets to study \mname{} :
\begin{itemize}
    \item FB15k-237~\cite{fb15k-237} consists of 14,505 entities, 237 relations, and 272,115 triplets when not considering the inverse of relations. In the process of obtaining relation-induced context information from the query graph, we utilize 544,230 triplets considering the inverse of relations.
    \item NELL~\cite{nell995} comprises 63,361 entities, 400 relations, and 114,213 edges without considering the inverse of relations. We utilize 228,426 triplets considering the inverses when obtaining relation-induced context information from the query graph.
\end{itemize}


\subsection{Query Dataset}\label{query_dataset}
\begin{figure}[h]
\centering
\includegraphics[width=\linewidth]{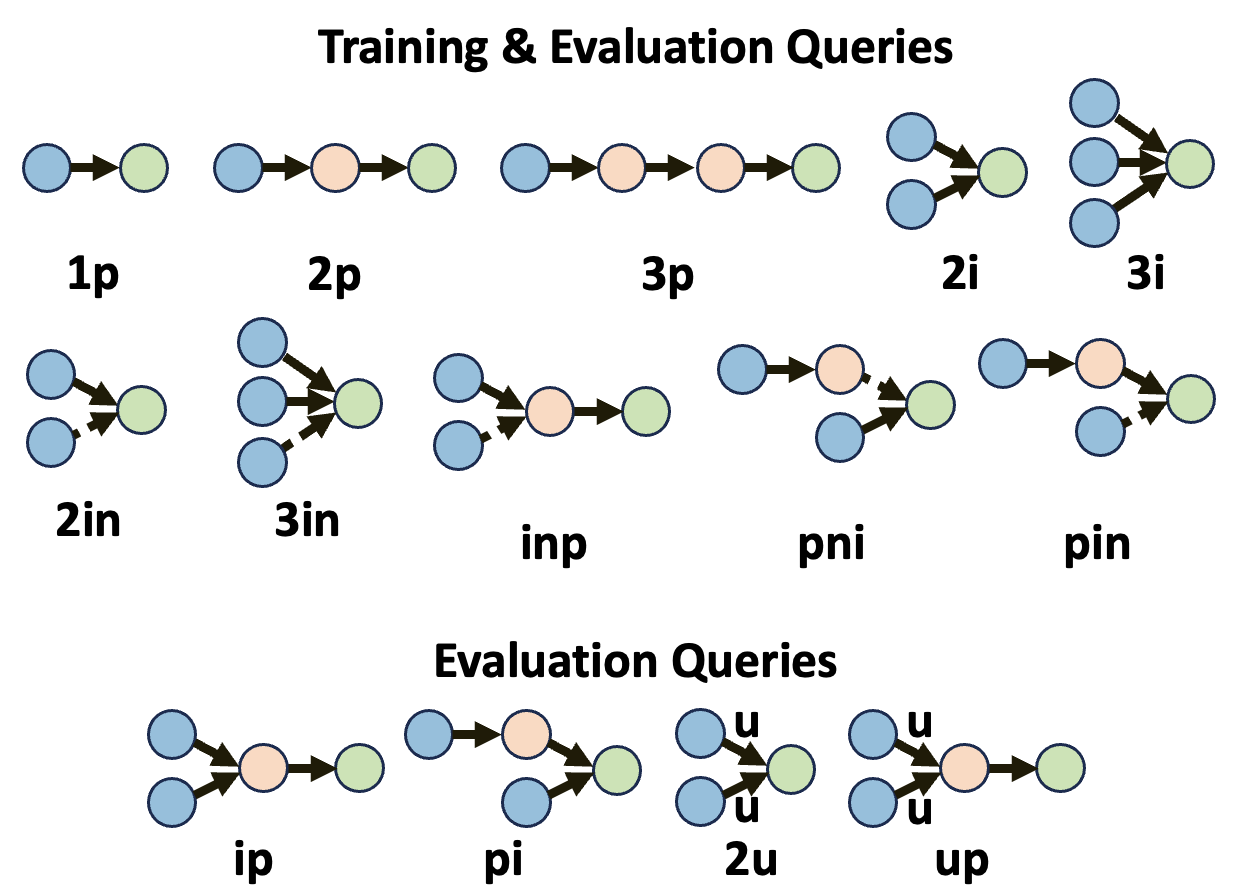}
\vspace{-4mm}
\caption{Query types used in experiments.}
\label{query_types}
\end{figure}

The query data employed for the experiment comprises 14 types as illustrated in Figure~\ref{query_types} with 4 of them (i.e., \textit{ip}, \textit{pi}, \textit{2u}, and \textit{up}) exclusively utilized for evaluation purposes. The statistics of the query data are provided in Table~\ref{query_statistics}.

\begin{table}[ht]
\caption{The Statistics of Query Datasets. Neg. means queries with negation operators.}
\label{query_statistics}
\resizebox{\columnwidth}{!}{
    \begin{tabular}{c|c|c|c|c|c|c}
    \hline
    \multirow{2}{*}{Dataset} &
    \multicolumn{2}{c|}{Training} &
    \multicolumn{2}{c|}{Validation} &
    \multicolumn{2}{c}{Test} \\ \cline{2-7}
     & w/o Neg. & Neg. & 1p & others & 1p & others \\
    \hline
    FB15k-237 & 273,710 & 27,371 & 59,078 & 8,000 & 66,990 & 8,000 \\
    \hline
    NELL & 149,689 & 14,968 & 20,094 & 5,000 & 22,804 & 5,000 \\
    \hline
    \end{tabular}
}
\end{table}

\subsection{Base Reasoning Models}

Since our approach is model-agnostic, it can be applied to any query embedding-based multi-hop reasoning model. Therefore, we evaluate the effectiveness of our approach on the Q2B~\cite{q2b}, BetaE~\cite{betae}, and ConE~\cite{cone} models for conjunctive query types with path lengths 1, 2, and 3 (\textit{1p, 2p, 3p, 2i, 3i, 2in, 3in, inp, pni, pin, ip, pi, 2u, and up}). Note that evaluations involving negations (\textit{2in, 3in, inp, pni, and pin}) are excluded on Q2B due to their limitations of the inability to address negations.

\section{Computation Graph} \label{computation_graph}
The computation graph shows a computationally feasible form of query graph using logical operators. 
Each query graph can be mapped into its corresponding computation graph~\cite{q2b}, where each atomic formula is represented with relation projection, merged by intersection, and transformed negation by complement.
The computation graph effectively demonstrates the computational procedure to resolve the query. In a computation graph, each node represents an embedding or a distribution over a set of entities in the KG, and each edge signifies a logical transformation (e.g., relational projection, intersection/union/negation operators) of this distribution. The computation graph for a FOL query resembles a tree. The root node of the computation graph represents the answer (or target) variable, with one or more anchor nodes provided by the FOL. Embeddings of entities and transformation operators are initialized; embeddings of anchor nodes are then input into the connected neural network of logical operators in a serial manner to obtain the final embeddings for the answer variable, thereby creating a query embedding. During training, models ensure the proximity of query embeddings to the ground truth. In the prediction stage, entities close to the query embedding are utilized for prediction.
We follow the original implementations of the base reasoning models, such as Q2B, BetaE, and ConE. Four logical operators are used to express the first-order logic queries: relation projection operator, intersection operator, union operator, and negation operator. 
\begin{itemize}
    \item \textbf{Relation projection operator.}\, For a set of entities $S \subseteq \mathcal{V}$, and a relation type $r \in \mathcal{R}$, the relation projection operator produces all the neighboring entities $\cup_{v \in S}N(v, r)$, where $N(v,r) \equiv \{v' \in \mathcal{V} \,|\, r(v, v') = True\}.$

    \item \textbf{Intersection operator.}\, For a set of entity sets, $\{S_1, S_2, ..., S_n\}$ of size $n$, the intersection operator outputs $\cap_{i=1}^{n}S_i$.

    \item \textbf{Union operator.}\, Given a set of entity sets, $\{S_1, S_2, ..., S_n\}$ of size $n$, the union operator produces $\cup_{i=1}^{n}S_i$.

    \item \textbf{Negation operator.}\, The negation operator outputs a complement set, $\overline{S} \equiv \mathcal{V} \backslash S $, of a set of entities $S \subseteq \mathcal{V}$.

\end{itemize}


\section{Case Study}





\begin{figure}[h]
\centering
\includegraphics[width=\linewidth]{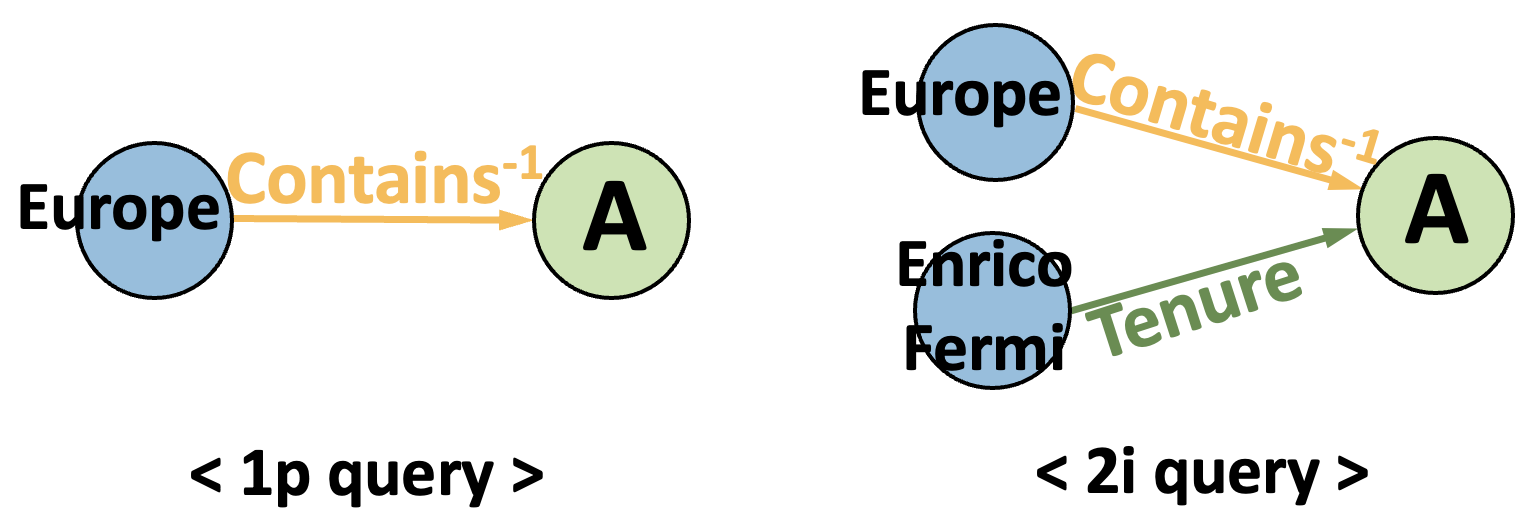}
\vspace{-4mm}
\caption{1p query and 2i query.}
\label{1p_2i}
\end{figure}

\begin{figure}[t]
    \centering
    \begin{subfigure}{\linewidth} 
        \centering
        \includegraphics[width=\linewidth]{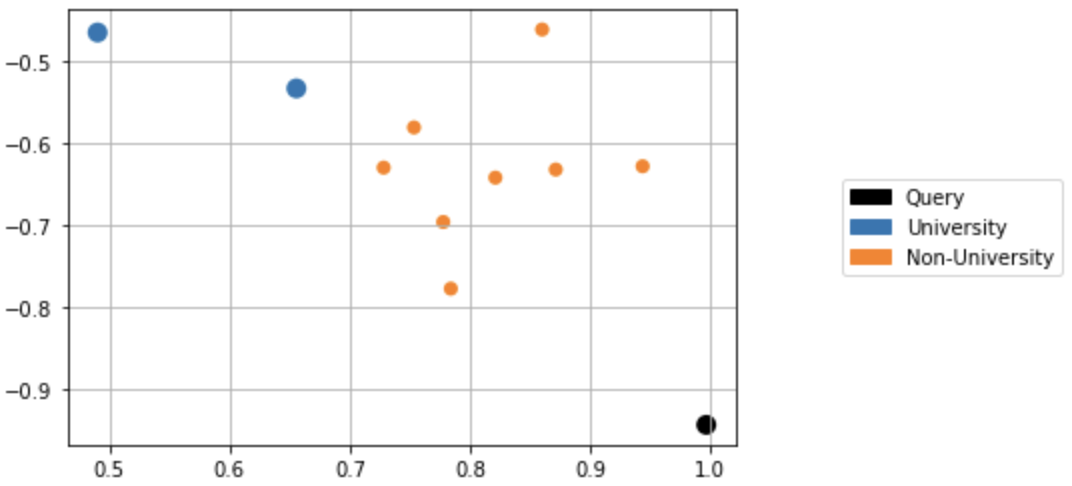}
        \caption{Result from Q2B}
        \label{case_q2b}
    \end{subfigure} 
    \begin{subfigure}{\linewidth} 
        \centering
        \includegraphics[width=\linewidth]{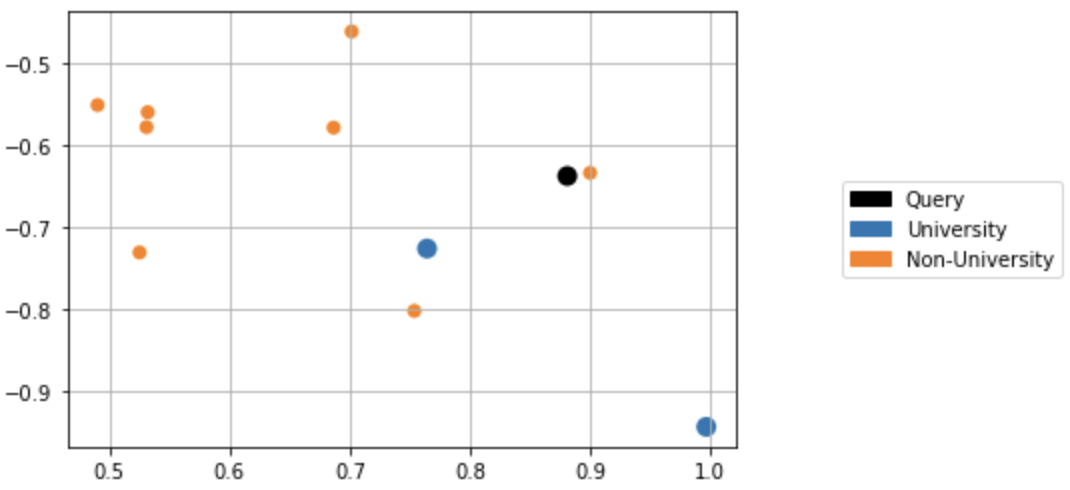}
        \caption{Result from Q2B + \both{}}
        \label{case_hprs}
    \end{subfigure}

    \caption{The embedding results of two models for a branch $Contains^{-1}(Europe, A)$ in the \textit{2i} query.}
    \label{case_study}
\end{figure}

To investigate whether our model reflects the context of the query graph in its embeddings, we conduct a case study using the 1p query $Contains^{-1}(Europe, A)$ and the 2i query $Contains^{-1}(Europe, A) \wedge Tenure(Enrico, A)$ (Figure~\ref{1p_2i}). When employing the conventional query embedding (Q2B) approach, it is not possible to capture the overall context of the query graph, thus resulting in the same query embedding for the $Contains^{-1}(Europe, A)$ in \textit{1p} and \textit{2i} queries. On the contrary, when applying our model, we observe the difference between $Contains^{-1}(Europe, A)$ embedding of \textit{1p} and \textit{2i} query indicating the incorporation of the query graph's context into the embeddings. \\
As shown in Figure~\ref{case_q2b}, through Q2B, it is impossible to obtain embedding that reflects the query graph context. Consequently, around the embedding of the $Contains^{-1}(Europe, A)$ branch (black), entities unrelated to the branch $Tenure(Enrico Fermi, A)$, which are not in the domain of \textit{University} are mapped in close proximity. Conversely, upon applying our model, as depicted in Figure~\ref{case_hprs}, an adaptation based on the context of the query graph becomes apparent. The embedding of the $Contains^{-1}(Europe, A)$ branch (black) is influenced by the $Tenure(Enrico Fermi, A)$ branch, leading to entities associated with \textit{University} close to the query embedding. This case shows the capacity to obtain refined representations by incorporating query context, yielding more accurate answers.



\section{Variance Loss}\label{appendix_variance}
Our approach ensures that the centers of query embedding, whether they are central points within the coordinate system, the averages of probability distributions, or the central axis of a cone in a polar coordinate, experience nuanced adjustments based on the contextual information inherent to the query. This leads to an enhanced representation by integrating both the overall structural and the relation-induced context within the query graph.
However, the probability-based query embedding approach, such as BetaE ~\cite{betae}, employs parameters ($\alpha$, $\beta$) of multiple beta distributions as query embeddings. In this method, when the embedding changes, it not only affects the centers but also alters the variances. Consequently, to mitigate unintended shifts in variance hindering the convergence of a model, we introduce a variance loss to BetaE. 
The adjusted embedding, $\mathbf{q}_A'$, does not deviate significantly from the offset of the query embedding composed of the parameters learned by the BetaE, $\mathbf{q}_A$, as follows:
\begin{align}
    \mathcal{L}_{var} = ||Var(\mathbf{q}_A)-Var(\mathbf{q'}_A)||_2,
\label{var_loss}
\end{align}where $Var(\cdot)$ returns a vector of variances of beta distributions consisting of Beta embedding. 
The loss considering Equation (\ref{var_loss}) is as follows:
\begin{align}\label{total_loss}
    \mathcal{L} = \mathcal{L}_{qe} + \lambda \mathcal{L}_{var},
\end{align}where the $\lambda$ is a hyper-parameter that determines the weighting of the variance loss. Details about Equation~(\ref{var_loss}) are provided in Appendix~\ref{appendix}.
Detailed formula of Equation~(\ref{var_loss}) can be written as follow:
\begin{align}
    & \textbf{q}_A = [(\alpha_1, \beta_1), ..., (\alpha_d, \beta_d)], \\
    & \textbf{q}_A' = [(\alpha'_1, \beta'_1), ..., (\alpha'_d, \beta'_d)].
\end{align}Here, $\textbf{q}_A$ and $\textbf{q}'_A$ are embedding learned by the BetaE, and adjust query embedding of answer node A of query graph, as described in Section~\ref{training} of the main text. $d$ is the dimension of BetaE, the number of beta distributions representing the query. When $Beta_i$ is an beta distribution parameterized by $\alpha_i$ and $\beta_i$, variance of $\textbf{q}_A$ and $\textbf{q}'_A$ are as follow:
\begin{align}
    & Var(\textbf{q}_A) = [\sigma(Beta_1), ..., \sigma(Beta_d)]\\           & Var(\textbf{q}'_A) = [\sigma(Beta'_1), ..., \sigma(Beta'_d)],
\end{align}where $\sigma(\cdot)$ is a variance of a probability distribution, and
\begin{align}
    &\sigma(Beta_i) = \frac{\alpha_i\beta_i}{(\alpha_i + \beta_i)^2(\alpha_i+\beta_i+1)},\\
    &\sigma(Beta'_i) = \frac{\alpha'_i\beta'_i}{(\alpha'_i+\beta'_i)^2(\alpha'_i+\beta'_i+1)}.
\end{align} Then, we add the L2 norm of variance difference between the query embedding constructed with pre-trained BetaE and the adjusted query embedding to the query-answering loss as described in Equation~(\ref{total_loss}) with coefficient $\lambda$.


\section{Inference Time Comparison}
We evaluate the inference times of both the baseline and the baseline with our methodology applied, showcasing the outcomes in Table~\ref{inference_time}. 

\begin{table}
\resizebox{\columnwidth}{!}{
\centering
\begin{tabular}{cccc}

\hline
\multirow{2}{*}{Model} & \multicolumn{3}{c}{Inference Time ($m$s)} \\
& \textit{3p} & \quad \textit{3i} & \textit{ip} \\
\hline
Q2B & 2.0 ($\pm$ 1.6) & 2.3 ($\pm$ 1.3) & 2.6 ($\pm$ 1.9) \\
Q2B + \strc{} &  3.3 ($\pm$ 2.3) &  2.9 ($\pm$ 2.5) & 3.8 ($\pm$ 2.7) \\
Q2B + \local{} & 5.4 ($\pm$ 4.3) & 3.4 ($\pm$ 2.4) & 3.6 ($\pm$ 1.5) \\
Q2B + \both{} & 9.6 ($\pm$ 6.9) & 4.6 ($\pm$ 3.2) & 6.5 ($\pm$ 4.7) \\
\hline

\end{tabular}
}
\caption{Averages and standard deviations of inference time on 1,000 queries involving complex query types (\textit{3p}, \textit{3i}, \textit{ip}). \strc{} denotes the model with position, role, and type embedding. \local{} indicates the model with relation-induced embedding.}
\label{inference_time}
\end{table}

\section{Further Experiment}
In addition to experiments conducted on Q2B, BetaE, and ConE, we further performed experiments applying our CaQR to the FuzzQE~\cite{fuzzqe} using the FB15k-237 dataset. The experimental results are on the Table~\ref{caqr_fuzzqe}. We observe performance improvements even for FuzzQE.

\begin{table}
    \centering
    \caption{MRR results (\%) of applying \mname{} to FuzzQE on FB15k-237 dataset. Avg\_P, Avg\_N, and Avg represent the average results for query types without negation, query types with negation, and all query types, respectively.}
    \label{caqr_fuzzqe}
    \scalebox{0.87}{
    \begin{tabular}{c|c|c|c|c}
    \hline
    \hline
        Model & Avg\_P & Avg\_N & Avg & Imp(\%) \\ \hline
        FuzzQE & 21.19 & 6.91 & 16.09 & - \\ \hline
        FuzzQE+\mname{} & 23.10 & 7.13 & 17.39 & 8.1 \\ \hline
    \end{tabular}}
\end{table}

\end{document}